\documentclass[conference]{IEEEtran}
\IEEEoverridecommandlockouts
\usepackage{amsmath,amssymb,amsfonts}
\usepackage{algorithmic}
\usepackage{graphicx}
\usepackage{textcomp}
\usepackage{xcolor}
\usepackage{graphicx}
\usepackage[square, numbers]{natbib}
\usepackage{wrapfig}

\usepackage[utf8]{inputenc} 
\usepackage[T1]{fontenc}    
\usepackage{
hyperref}       
\usepackage{url}            
\usepackage{booktabs}       
\usepackage{amsfonts}       
\usepackage{nicefrac}       
\usepackage{microtype}      
\usepackage{xcolor}         

\def\BibTeX{{\rm B\kern-.05em{\sc i\kern-.025em b}\kern-.08em
    T\kern-.1667em\lower.7ex\hbox{E}\kern-.125emX}}
\begin{document}

\title{Integrating MedCLIP and Cross-Modal Fusion for Automatic Radiology Report Generation}

\author{\IEEEauthorblockN{Qianhao Han}
\IEEEauthorblockA{
\textit{The Bishop Strachan School}\\
Toronto, Canada \\
qianhaoh27@bss.on.ca}
\and
\IEEEauthorblockN{Junyi Liu}
\IEEEauthorblockA{\textit{Baidu Inc.} \\
\textit{Beihang University}\\
Beijing, China \\
liujunyi671@buaa.edu.cn}
\and
\IEEEauthorblockN{Zengchang Qin}
\IEEEauthorblockA{\textit{Institute of Future Education} \\
\textit{TechArena Canada Inc.}\\
\textit{Beihang University.}
Beijing, China \\
zcqin@buaa.edu.cn}
\and
\IEEEauthorblockN{Zheng Zheng}
\IEEEauthorblockA{\textit{College of Engineering} \\
\textit{Northeastern University}\\
Boston, USA \\
zh.zheng@northeastern.edu}
}

\maketitle

\begin{abstract}
  Automating radiology report generation can significantly reduce the workload of radiologists and enhance the accuracy, consistency, and efficiency of clinical documentation.  
  We propose a novel cross-modal framework that uses MedCLIP as both a vision extractor and a retrieval mechanism to improve the process of medical report generation. 
  By extracting retrieved report features and image features through an attention-based extract module, and integrating them with a fusion module, our method improves the coherence and clinical relevance of generated reports. 
  Experimental results on the widely used IU-Xray dataset demonstrate the effectiveness of our approach, showing improvements over commonly used methods in both report quality and relevance. 
  Additionally, ablation studies provide further validation of the framework, highlighting the importance of accurate report retrieval and feature integration in generating comprehensive medical reports. 
\end{abstract}


\begin{figure*}[htbp]
  \centering
  \includegraphics[width=7.2in, trim=20 10 0 10]{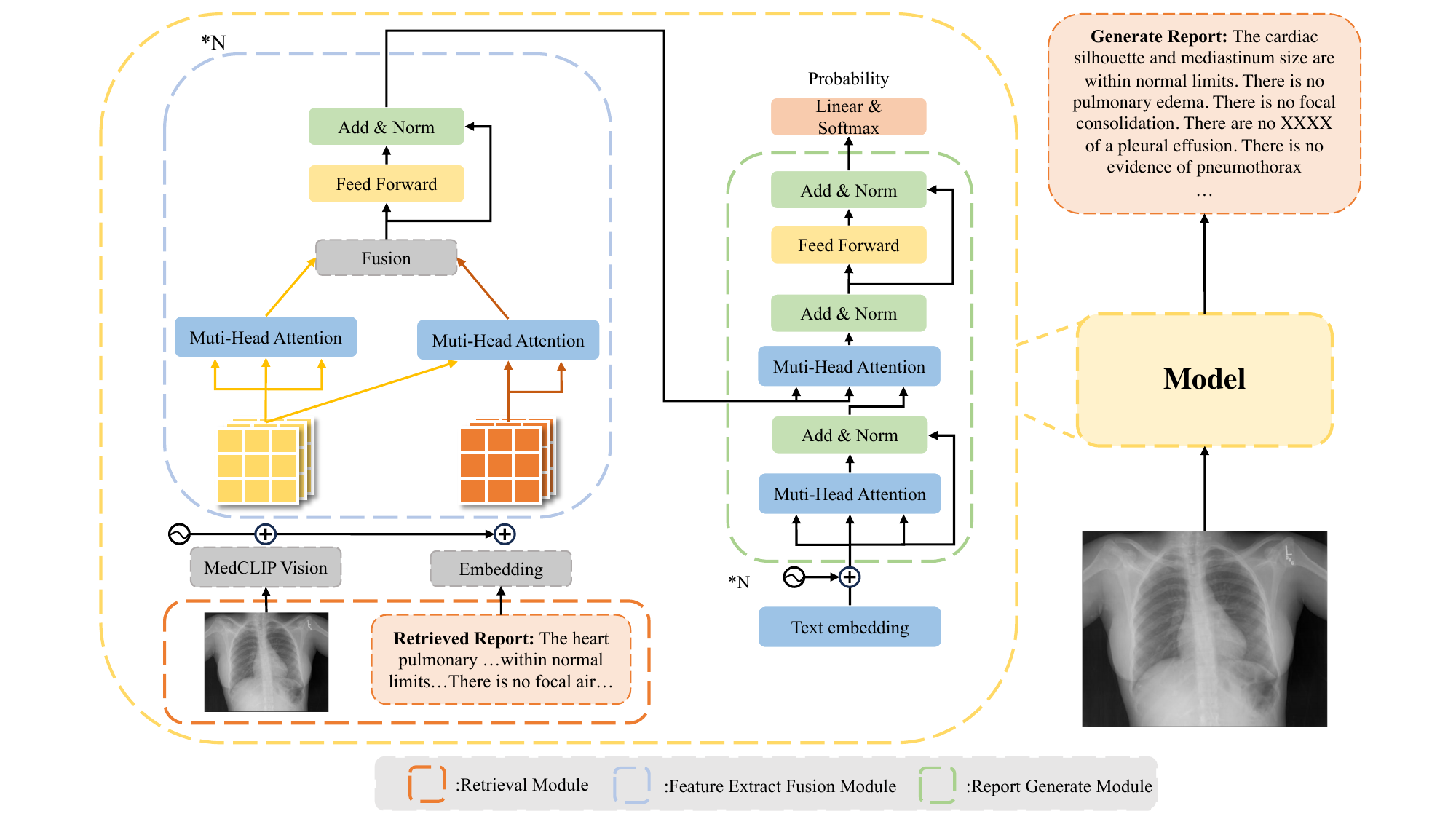}
  \caption{\footnotesize The framework of the proposed method}
  \label{fig:CLIPR2G}
  \vspace{-7pt}
\end{figure*}

\section{Introduction}
Medical imaging is integral to clinical diagnosis, providing crucial insights that guide subsequent treatments. 
Interpreting radiology images (e.g., X-rays, CTs, and MRIs) and writing diagnostic reports are critical tasks in clinical practice that require substantial manual effort and expertise.
However, the manual generation of these reports by radiologists is time-consuming and labour-intensive. It could also be prone to bias, especially for less experienced radiologists. 
Automating the report-writing process can significantly alleviate the workload of radiologists, allowing them to focus on more critical tasks.  Recent advancements in artificial intelligence, particularly in cross-modal large models, have shown great promise in addressing this challenge in other tasks. 

Related research has been reported on designing cross-modal models for medical images and informative reports, mimicking the expertise of human radiologists. Among these efforts, MedCLIP\cite{zifeng2022medclip} stands out as an improvement upon CLIP\cite{radford2021clip} by using contrastive learning to handle unpaired medical images and text. This approach leverages a large corpus of unpaired data, enabling the model to more effectively bridge the gap between visual and textual information, thereby enhancing its ability to perform accurate medical imaging tasks. Based on its promising results, we propose a novel cross-modal framework using MedCLIP as a vision extractor and retrieval mechanism to enhance report generation. Our method extracts report and image features through an attention-based module, and integrates them with a fusion module. The experimental results demonstrate that our method improves the coherence and clinical relevance of the generated reports.

\section{Related Work}
Numerous approaches (\cite{anderson2018bottom}, \cite{wang2018tienet}, \cite{cornia2020meshed}, \cite{zhou2020more}, \cite{liu2019clinically}, \cite{jing2019show}) have been proposed to tackle image captioning, which involves automatically generating short descriptions for natural images. Most existing works rely on a conventional encoder-decoder architecture, integrating convolutional neural networks (CNNs) for image feature extraction with recurrent neural networks (RNNs) or non-recurrent networks for description generation. Despite their success, these methods struggle to generate detailed, long-form reports due to the limited ability to capture the intricate characteristics of radiology images.

Recently, vision-language models have become crucial in medical imaging, with many relying on transformer-based architectures to generate descriptive and diagnostic reports. This is largely due to transformers' ability to handle long-range dependencies and process sequential data efficiently, making them well-suited for the complex task of interpreting medical images and generating corresponding textual reports.
\citet{zhang2020when} utilize knowledge graphs to improve the accuracy and contextual relevance of generated reports.
\citet{chen2020generating} develop a multi-modal transformer model that integrates memory mechanisms to capture long-term dependencies and context.
\citet{chen2021cross} employ cross-modal memory networks to align visual and textual data, improving report accuracy by maintaining and leveraging important information across different modalities.
With increasing model size, research has shifted to vision-language pre-training models. CLIP\cite{radford2021clip} uses a contrastive learning framework to align visual and textual representations, performing well on various tasks without task-specific fine-tuning. 

To address the limitations in the alignment between visual and textual representation in medical imaging,
MedCLIP\cite{zifeng2022medclip} is proposed to use a contrastive learning framework, which significantly enhances image-text characterization in the medical domain.
\citet{zhao2024chatcad} employs large language models (LLM) for interactive computer-aided diagnosis, significantly improving diagnostic accuracy and decision-making through real-time, context-aware interactions between the model and medical practitioners.

\section{The Proposed Approach} \label{approach}

Our proposed framework employs an encoder-decoder architecture to automatically generate radiology reports input radiographic images (as shown in Figure \ref{fig:CLIPR2G}). The architecture comprises three core modules: a similar report retrieval module, a feature extraction fusion module, and a report generation module. We denote a radiographic image as $I$ and its corresponding radiology report as $R = \{y_1, y_2, \ldots, y_n\}$, where $y_i$ represents each word token in the generated report, and $n$ is the length of the report. The input to our framework is the radiograph $I$, and the output is the report $R$. Our approach is formulated as follows:

\begin{equation}
    R_r = MR(I)
\end{equation}
\begin{equation}
    F_e = MoF(MedCLIP(I),Embedding(R))
\end{equation}
\begin{equation}
    R = Transformer(F_e)
\end{equation}


Specifically, $MR()$ refers to the process of using MedCLIP\cite{zifeng2022medclip} to retrieve a report that is most similar to the input radiograph $I$. $MoF()$, which stands for Mixture of Feature modules, is responsible for combining the visual features extracted from the current radiograph with the textual features from the retrieved report. $Transformer()$ represents the report generation phase, where a classical end-to-end Transformer architecture is used to produce the final radiology report.
Additionally, $MedCLIP()$ denotes using MedCLIP model to extract detailed image feature. $Embedding()$ refers to the representation of the report text as embeddings.

\begin{table*}[h]
    \centering
    \caption{Results compared with previous work}
    \begin{tabular}{c|cccccc}
    \toprule
    \textbf{Method} & BLEU-1 & BLEU-2 & BLEU-3 & BLEU-4 & ROUGE-L & METEOR \\
    \midrule
    CoAttn & 0.455 & 0.288 & 0.205 & 0.154 & 0.369 & - \\
    R2Gen & 0.470 & 0.304 & 0.219 & 0.165 & 0.371 & 0.187 \\
    R2GenCMN & 0.470 & 0.304 & 0.222 & 0.179 & 0.358 & 0.190 \\
    Ours  & 0.469 & 0.307 & 0.223 & 0.172 & 0.365 & 0.194 \\
    \bottomrule
    \end{tabular}
    \label{tab:comparison tab}
\end{table*}

\begin{table*}[h]
    \centering
     \caption{Results of ablation study}
    \begin{tabular}{c|cccccc}
    \toprule
    \textbf{Method} & BLEU-1 & BLEU-2 & BLEU-3 & BLEU-4 & ROUGE-L & METEOR \\
    \midrule
    Baseline\cite{chen2021cross} & 0.396 & 0.254 & 0.179 & 0.135 & 0.342 & 0.164 \\
    +MedCLIP  & 0.481 & 0.313 & 0.229 & 0.177 & 0.369 & 0.196 \\
    +MedCLIP+Retrieval  & 0.469 & 0.307 & 0.223 & 0.172 & 0.365 & 0.194 \\
    \bottomrule
    \end{tabular}
    \label{tab:abalation tab}
\end{table*}

\subsection{Similar Report Retrieval Module} \label{retrieval}
Inspired by radiologists referring to existing cases when writing reports, we employ a retrieval method to find the most similar reports from a report database to assist in generating new reports. 
ResNet-50\cite{he2016resnet50} is commonly used to extract image features. The most similar images to the input image were then retrieved, and the corresponding reports of these images were used as the retrieved similar reports. However, since ResNet-50 is pre-trained on ImageNet\cite{deng2009imagenet}, it lacks the ability to effectively capture medical-specific features.

To address this, we use MedCLIP\cite{zifeng2022medclip}, which aligns medical image and text features through contrastive learning to extract more effective visual representations. We represent the process as:
\begin{equation}
    F_{I_i} = MedCLIP(I_i)
\end{equation}
where $F_{I_i}$ represents a radiograph feature. We extract all image features from the Retrieval Dataset (denoted as $D_r$) and calculate the cosine similarity between $F_{I_i}$ and each $F_{I_j} \in D_r$. The report corresponding to the image with the highest similarity is selected as the retrieved report. This process can be represented as:
\begin{equation}
    R_{r_i} = \max(similarity(F_{I_i},F_{I_j}))F_{I_j} \in D_r
\end{equation}

\subsection{Feature Extraction Fusion Module}
After the retrieval module, the input consists of an image \( I_i \) and a retrieval report \( R_{r_i} \). Given that these two modalities are distinct, we employ an attention mechanism to effectively fuse information from both, facilitating report generation. Following the method described in Section~\ref{retrieval}, we utilize MedCLIP to extract image features and employ text embeddings to capture the report features.
\begin{equation}
    F_{I_i} = MedCLIP(I_i) 
\end{equation}
\begin{equation}
    F_{R_{r_i}} = Embedding(R_{r_i})
\end{equation}
After acquiring features from both modalities, our initial step involves extracting information directly from the image, given its importance as the primary information source. To accomplish this, we employ a self-attention mechanism to focus on:
\begin{equation}
    Att(Q,K,V) = Softmax(QK^{T}/\sqrt{d})V
\end{equation}
\begin{equation}
    I_{I} = Att(F_{I_i},F_{I_i},F_{I_i})
\end{equation}
The function \( Att() \) represents the attention module in the Transformer architecture. We employ the image feature as the query, the retrieval report feature as the key, and the value to extract relevant retrieval information pertaining to the image. This process utilizes a cross-attention mechanism:
\begin{equation}
    I_{R} = Att(F_{I_i},F_{R_{r_i}},F_{R_{r_i}})
\end{equation}
We employed a weighted mixing method to fuse information between the image and the report, which can be formulated as:
\begin{equation}
    F = I_{I} + W \cdot I_{R}
\end{equation}
where \( W \) is a learnable parameter and \( F \) is the fusion feature of the image and the report. We utilize this fusion feature to generate the final report.

\subsection{Report Generation Module}
We adopt the classic encoder-decoder architecture to generate reports. The encoder layer adopts the fusion modules with the Feed Forward Network, which can be formulated as:
\begin{equation}
    F = Fusion(I,R_{r_I})
\end{equation}
\begin{equation}
    F = F + Drop(LN(F))
\end{equation}
\begin{equation}
    E_o = F + Drop(LN(FFN(F)))
\end{equation}
\begin{equation}
    FFN(F) = W_2 \cdot (ReLU(W_1*F + B_1)) + B_2)
\end{equation}
where $Drop()$ means Dropout and $LN()$ means Layernorm. For the decoder, we use the classical Transformer decoder. Given the ground truth report $R^*= \{y_1^*,y_2^*,...,y_n^*\}$, we can train the model by minimizing the cross-entropy loss:
\begin{equation}
    \mathcal{L}_{CE}(\theta) = -\sum_{i=1}^{n}\log p_\theta(y_i = y_i^*|y_{1:i-1}^*,I)
\end{equation}

\begin{figure*}[htbp]
  \centering
  \includegraphics[width=6.5in, trim=20 10 0 10]{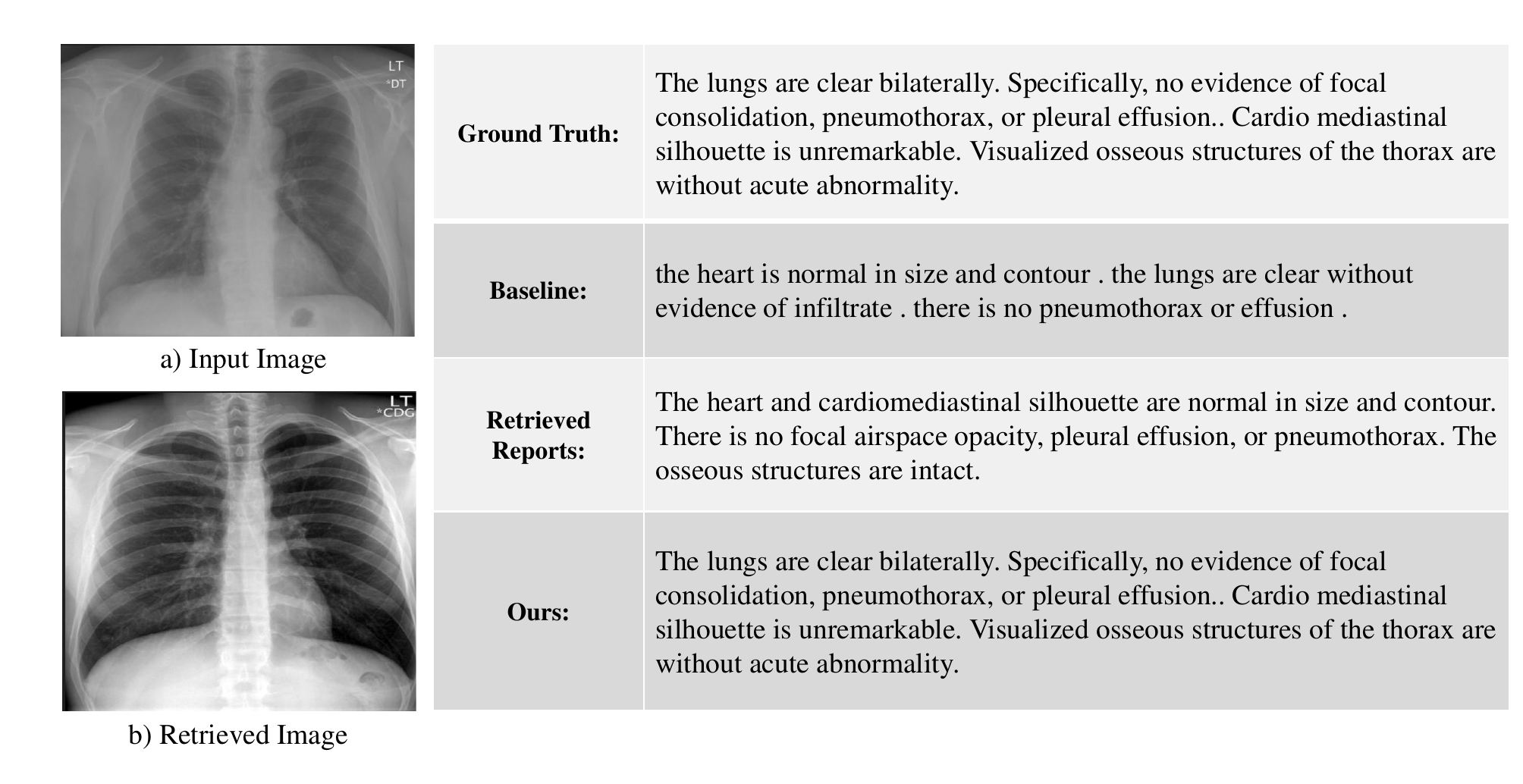}
  \caption{\footnotesize The case study of the proposed method}
  \label{fig:casestudy}
  \vspace{-7pt}
\end{figure*}
\section{Experimental Studies} \label{experiment}

\subsection{Dataset.} 
We perform our experiments using the publicly available IU-Xray benchmark dataset\cite{demner2016preparing}, which consists of 3,955 reports and 7,470 frontal and lateral radiology images. To ensure a fair comparison with previous methods, we followed the data preprocessing steps outlined in R2Gen \cite{chen2020generating} and divide the dataset into training, validation, and test sets with a ratio of 7:1:2. 

\subsection{Metrics and Settings.} 
To evaluate the performance of our model in generating radiology reports, we use the widely accepted Natural Language Generation metrics: BLEU\cite{papineni2002bleu}, METEOR\cite{banerjee2005meteor}, and ROUGE-L\cite{lin2004rouge}. During training, all images are resized to \(256 \times 256\) and randomly cropped to \(224 \times 224\). For testing, all images are resized directly to \(224 \times 224\). Following the approach in previous work \cite{chen2020generating}, we use both frontal and lateral X-ray images as input for IU-Xray by stacking them. Our encoder-decoder architecture is based on a randomly initialized Transformer with three layers. Reports are generated using the top retrieved report, and the model is trained for 100 epochs on the IU-Xray dataset with a beam search size of 3. The model is optimized using the Adam optimizer, with a learning rate of \(5 \times 10^{-5}\) for the visual extractor and \(5 \times 10^{-4}\) for the other modules.
All the experiments are conducted on an NVIDIA Tesla T4 with PyTorch 1.10. Code and dataset can be found at \href{https://github.com/QianhaoHan/Cross-Modality-Medical-Report-Generation}{Github}.


\subsection{Comparative Experiments.} 
We compare our approach with state-of-the-art methods CoAttn~\cite{jing-etal-2018-automatic}, R2GenCMN~\cite{chen-etal-2021-cross-modal} and R2Gen on the IU-Xray dataset, and the results, as shown in Table~\ref{tab:comparison tab}, , indicate that our method consistently outperforms R2GenCMN in several key metrics. Specifically, our model demonstrates improvements of $0.3\%$, $0.1\%$, $0.7\%$, and $0.4\%$ on BLEU-2, BLEU-3, ROUGE-L, and METEOR scores, respectively. The increase in BLEU-2, BLEU-3, and ROUGE scores is attributed to the use of retrieved reports, which contain expressions similar to the Ground Truth, thus improving the quality of the generated text. Meanwhile, the rise in the METEOR score results from MedCLIP capturing more medical visual features, allowing the generated report to focus on medical details. While there is a slight decrease in performance on the BLEU-1 and BLEU-4 scores, this trade-off highlights our model's ability to generate more accurate and fluent long-form reports, focusing on clinically relevant details.

\subsection{Ablation Study.}
Table \ref{tab:abalation tab} shows the main results of our ablation studies. The baseline reflects the performance of using the classic Transformer architecture from R2GenCMN. Using the MedCLIP vision extractor improves metrics comprehensively due to its ability to extract both visual and textual features through contrastive learning. As a result, the extracted visual features contain some textual information, leading to enhanced performance.
However, when retrieval modules are introduced, there is a slight decline in performance. This is likely due to errors or redundancies present in the retrieved reports, as MedCLIP already captures sufficient textual information. The addition of redundant data from retrieval can negatively impact the model's overall effectiveness.


\subsection{Case Study.}
To validate our results, we selected a challenging case from the IU-Xray dataset and compare our method with the Transformer baseline model. The results are shown in Figure~\ref{fig:casestudy}. While the Transformer model \cite{vaswani2017attention} is capable of capturing some features and generating corresponding reports, it exhibits two main issues. First, there are discrepancies between the generated reports and the original ground truth. For example, "no evidence of focal consolidation, pneumothorax, or pleural effusion" is simplified to "there is no pneumothorax or effusion." missing critical details. Second, certain important features, such as "cardiomediastinal silhouette" and "osseous structures," are omitted entirely. 

Our approach leverages MedCLIP for feature extraction, producing reports that provide more comprehensive information, address missing details, and capture fine-grained features. This results in reports that are more detailed and closely aligned with the ground truth, highlighting our method's advantage in generating accurate, complete medical reports.

\section{Conclusion} \label{conclusion}
In this paper, we propose a novel cross-modal framework for medical report generation using MedCLIP as both a visual extractor and a retriever. A fusion module integrates image and retrieved report features to learn cross-modal information. Experimental results on public benchmarks demonstrate the framework's effectiveness.  Ablation studies validate the effectiveness of the proposed components. We identify potential errors or redundancies in the retrieved reports, suggesting that future research could focus on improving the accuracy of retrieved reports and implementing filtering mechanisms.

\bibliographystyle{plainnat}
\bibliography{references}

\end{document}